\pdfoutput=1

\documentclass[11pt]{article}

\usepackage[final]{naacl2021}
\usepackage{times}
\usepackage{latexsym}

\usepackage[T1]{fontenc}

\usepackage[utf8]{inputenc}

\usepackage{microtype}

\usepackage{amsmath}
\usepackage{amssymb}
\usepackage{array}
\usepackage{pifont}
\usepackage{microtype}
\usepackage{tabularx}
\usepackage{adjustbox}
\DeclareMathOperator*{\argmax}{arg\,max}
\usepackage{multirow}
\usepackage{booktabs,subcaption,amsfonts,dcolumn}
\usepackage{bbm}

\mathchardef\ordinarycolon\mathcode`\:
\mathcode`\:=\string"8000
\begingroup \catcode`\:=\active
  \gdef:{\mathrel{\mathop\ordinarycolon}}
\endgroup

\usepackage[compact]{titlesec}
\titlespacing{\section}{0pt}{2ex}{1ex}
\titlespacing{\subsection}{0pt}{1ex}{1ex}

\newcommand{\cmark}{\ding{51}}
\newcommand{\xmark}{\ding{55}}

\newcommand\ti[1]{\textit{#1}}

\newcommand\tf[1]{\textbf{#1}}

\newcommand{\xv}{\texttt{[X]}}
\newcommand{\vv}{\texttt{[V]}}
\newcommand{\tv}{\texttt{[T]}}
\newcommand{\mask}{\texttt{[MASK]}}

\newcommand{\lama}{\textsc{LAMA}}
\newcommand{\lamauhn}{\textsc{LAMA-UHN}}
\newcommand{\lpaqa}{\textsc{LPAQA}}
\newcommand{\autop}{\textsc{AutoPrompt}}
\newcommand{\ours}{\textsc{OptiPrompt}}
\newcommand{\finetuning}{{Fine-tuning}}

\newcommand{\lamanumber}{42.2\%}
\newcommand{\ournumber}{48.6\%}
\newcommand{\uhnnumber}{31.3\%}
\newcommand{\ouruhnnumber}{38.4\%}

\title{Factual Probing Is [MASK]: Learning vs. Learning to Recall}

\author{Zexuan Zhong$^*$ \quad Dan Friedman$^*$ \quad Danqi Chen \\
Department of Computer Science \\ Princeton University  \\
\texttt{\{zzhong, dfriedman, danqic\}@cs.princeton.edu }
}

\begin{document}
\maketitle
\renewcommand{\thefootnote}{\fnsymbol{footnote}}
\footnotetext[1]{The first two authors contributed equally.}
\renewcommand{\thefootnote}{\arabic{footnote}}


\begin{abstract}
\citet{petroni2019language} demonstrated that it is possible to retrieve world facts from a pre-trained language model by expressing them as cloze-style prompts and interpret the model's prediction accuracy as a lower bound on the amount of factual information it encodes.
Subsequent work has  attempted to tighten the estimate by searching for better prompts, using a disjoint set of facts as training data.
In this work, we make two complementary contributions to better understand these factual probing techniques.
First, we propose {\ours}, a novel and efficient method which directly optimizes in continuous embedding space.
We find this simple method is able to predict an additional 6.4\% of facts in the LAMA benchmark.
Second, we raise a more important question:
Can we really interpret these probing results as a lower bound? Is it possible that these prompt-search methods learn from the training data too?
 We find, somewhat surprisingly, that the training data used by these methods contains certain regularities of the underlying fact distribution, and all the existing prompt methods, including ours, are able to exploit them for better fact prediction.
We conduct a set of control experiments to disentangle ``learning'' from ``learning to recall'', providing a more detailed picture of what different prompts can reveal about pre-trained language models.\footnote{The code is publicly available at \url{https://github.com/princeton-nlp/OptiPrompt}.}

\end{abstract}


\section{Introduction}
\label{sec:intro}

\begin{figure}
    \centering
    \resizebox{0.9\columnwidth}{!}{%
    \includegraphics{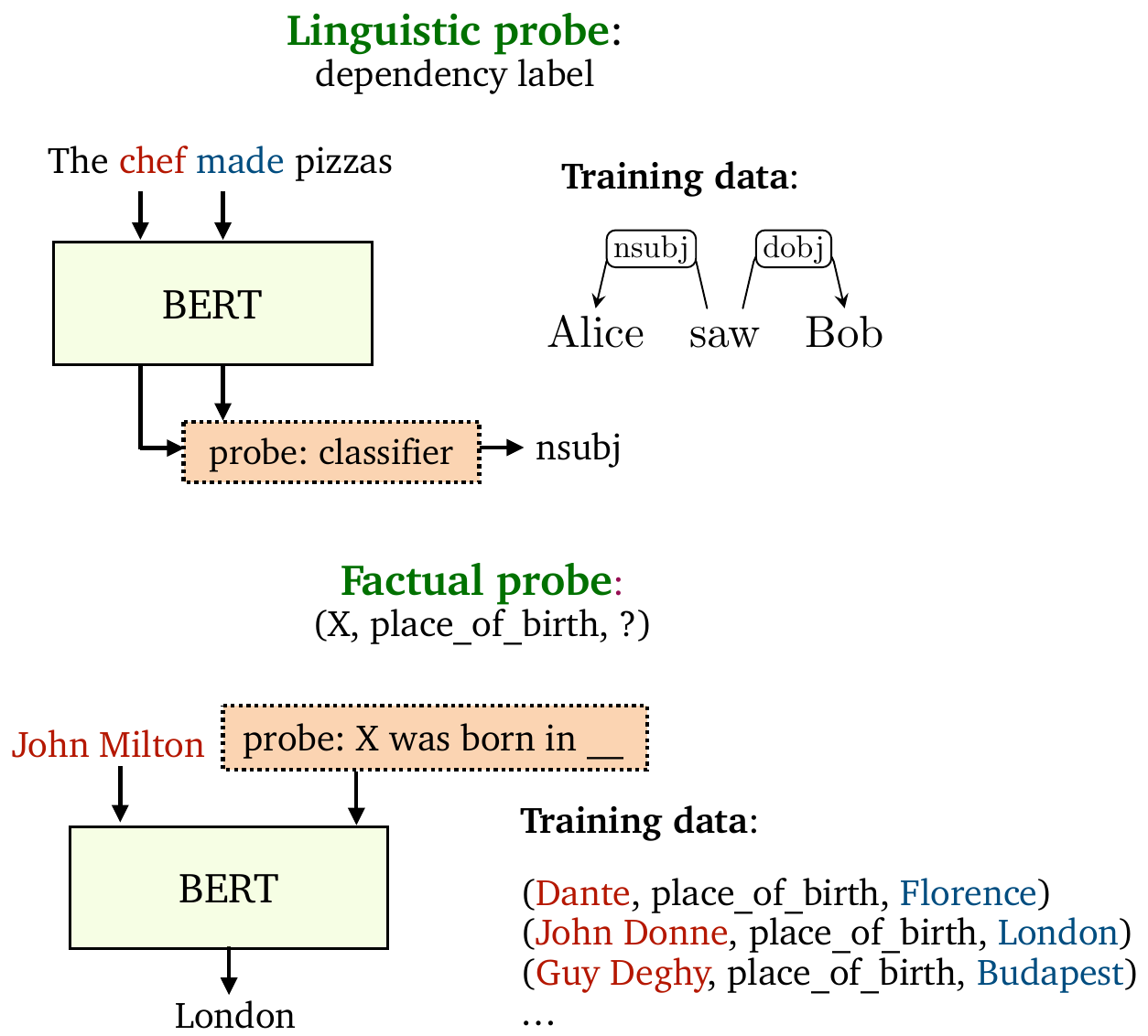}
    }
    \caption{A linguistic probe is trained to predict linguistic annotations given the representations returned by a language model, and evaluated on a held-out set of sentences.
    A factual probe is trained to predict an \ti{object} for a \ti{subject} and a \ti{relation} using a pre-trained language model, and evaluated on a held-out set of subject-object pairs that express the same relation.
    }
    \label{fig:two_probes}
\end{figure}



Pre-trained language models like BERT are optimized to predict the distribution of words in an Internet corpus \cite{devlin2018bert}. Naturally, this distribution encodes information about world facts. Recently, researchers have taken an interest in measuring how much factual information language models acquire from pre-training. \citet{petroni2019language} formally define this project in the {\lama} benchmark, which consists of (\ti{subject}, \ti{relation}, \ti{object}) triples along with human-written templates that express each relation. They show that BERT can predict objects given cloze-style prompts---for example, ``Dante was born in \mask''---and they present their result as a lower bound on the amount of factual information BERT encodes. Subsequent work has attempted to tighten this bound by finding better prompts. \citet{jiang2020can} use text mining and paraphrasing to find a set of candidates and select the prompts that lead to the highest accuracy on a training set. \citet{shin2020autoprompt} train a model to generate prompts automatically by searching for the sequence of tokens that maximizes expected likelihood of the gold object label. Both of these methods collect additional triples from Wikidata to use for tuning their prompts.



In this paper, we first take a natural next step in the search for better prompts: rather than confining our search space to discrete input tokens, we directly optimize in the input embedding space, finding the real-valued input vectors that are most effective at eliciting facts. We also find that initializing with manual prompts can provide a better starting point for the search process. Our approach, {\ours}, is simple and compute-efficient, and improves accuracy on the {\lama} benchmark from $\lamanumber$ to $\ournumber$, compared to previous discrete alternatives. On the more difficult LAMA-UHN split~\cite{poerner2019bert}, which filters out easy-to-guess entity names, {\ours} improves accuracy from {\uhnnumber} to {\ouruhnnumber}.



At the same time, we observe that prompts that are optimized on training data may exploit some regularities in the underlying distribution of facts.
How can we make sure our prompts are recovering information solely from the language model?
An analogous question has been explored recently in linguistic probing, which aims to explore the linguistic properties encoded in contextualized word representations~\cite{belinkov2017neural,tenney2019you,lin2019open}---for example, by seeing if a classifier can predict that ``\ti{chef}'' is the nominal subject of ``\ti{made}'' given the representations returned from a language model (Figure~\ref{fig:two_probes}). Recent work has attempted to disentangle the information encoded in the representations from the information learned by the probe \cite{hewitt2019designing,pimentel2020information,voita2020information,zhu2020information}.
However, this question has not been yet explored in factual probing, in part because it is assumed that there is no way to predict a knowledge fact simply from observing a non-overlapping set of facts about other entities.\footnote{In knowledge base completion or link prediction, researchers study how to predict a fact \ti{(Barack Obama, nationality, ?)} from other triples such as \ti{(Barack Obama, place\_of\_birth, Honolulu)} and \ti{(Honolulu, city\_of, USA)}. In knowledge probing, the underlying assumption is that one can't predict facts from the other facts of the \ti{same} relation.}
For example, learning that \ti{Dante} was born in \ti{Florence} should tell you nothing about the birthplace of \ti{John Donne}.

We analyze our training data and find that this assumption is not warranted. Even though the training data was collected independently of the {\lama} benchmark, there are sufficient regularities in the underlying distribution of Wikidata relations that a naive classifier fit to the training data can achieve surprisingly good performance.
Furthermore, our experiments reveal that all the data-driven prompt-search methods, including previous methods and our proposed {\ours}, are able to exploit this information to achieve better prediction accuracy. Given some training data, a good search algorithm can find prompts that recover a non-trivial number of ``facts'' from a neural network with randomly initialized parameters, exploiting both simple class statistics and higher order lexical regularities.

This finding makes it challenging to interpret relative accuracy scores on the knowledge probing task. We show how our control experiments allow us to form a more detailed understanding of the behavior of different probes. For example, by partitioning the test set into ``easy'' examples, which can be predicted by random controls, and ``hard'' examples, we can form some conclusions about which facts are less likely to have been learned from training data. {\ours} outperforms prior methods in both subsets, suggesting it is both better at learning from training data and better at eliciting facts from a language model. We conclude with suggestions for future work that might be less susceptible to the confounding effect of training data.

\begin{table*}
    \resizebox{2\columnwidth}{!}{%
    \begin{tabular}{llc}
        \toprule
        {\tf{Method}} & {\tf{Prompt}} & \tf{Data-driven?} \\
        \midrule
        \lama~\cite{petroni2019language} & {\xv} is {\mask} citizen & \xmark \\
        \midrule
        \lpaqa~\cite{jiang2020can} & {\xv} is a citizen of {\mask} & \cmark \\
        \autop~\cite{shin2020autoprompt} & {\xv} {\small m$^3$ badminton pieces internationally representing {\mask}} & \cmark \\
        \midrule
        \ours &  {\xv} $\vv_1$ $\vv_2$ $\vv_3$ $\vv_4$ $\vv_5$ \mask~ & {\cmark} \\
        \ours~(manual)& {\xv} $\vv_1:={\text{is}}$ {\mask}  $\vv_2:={\text{citizen}}$ & \cmark    \\
        \bottomrule
    \end{tabular}
    }
    \caption{Comparison of prompts for the relation \ti{country of citizenship}. {\xv} denotes the name of the subject and {\mask} is single-token object label to be predicted. In our {\ours} approach, we optimize a sequence of learned embeddings $\vv_i \in \mathbb{R}^d$ for each relation type. $\vv_i := w$ indicates that the vector is learned but initialized by the pre-trained embedding of word $w$ and {\ours} (manual) indicates that we use a manual prompt as initialization (see Section~\ref{sec:ours} for more details).
    }
    \label{tab:method_comparison}
\end{table*}


\section{Background: Prompting for Facts}
\label{sec:background}

\subsection{{\lama}}
\label{sec:lama}
The factual probing setting was introduced by the {\lama} benchmark~\cite{petroni2019language}, which is designed to measure the amount of factual information encoded in a pre-trained language model (LM).
In {\lama}, a fact is defined as a triple $\langle s, r, o \rangle$, where $s$ is a subject (e.g., \ti{Dante}), $r$ is a relation from a fixed set of relations $\mathcal{R}$ (e.g., \ti{place of birth}), and $o$ is an object (\ti{Florence}).
{\lama} facts are drawn from a number of sources, including Wikidata, ConceptNet~\cite{speer2012representing}, and SQuAD~\cite{rajpurkar2016squad}.
We follow recent factual probing work~\cite{jiang2020can,shin2020autoprompt} in focusing on the T-REx split \cite{trex2018elsahar}, which contains up to 1000 $\langle s, r, o \rangle$ triples for each of 41 Wikidata relation types.  The relation types are divided into three categories: 1-1 includes relations like \ti{capital of}; N-1 includes relations like \ti{place of birth}; and N-M includes relations like \ti{shares border with}. In the {\lama} evaluation, each relation is associated with a human-written prompt that contains a single {\mask} token---for example, ``{\xv} was born in {\mask}.''~To accommodate masked language models such as BERT, {\lama} is restricted to facts for which the object label is a single token in a predefined vocabulary $\mathcal{V}$.\footnote{Subject names are usually longer, with an average length of $3.7$ tokens using the BERT-base-cased vocabulary.}
Given a subject $s$, a relation prompt $t_r$, and a masked language model, we can identify the word $\hat{o} \in \mathcal{V}$ to which the LM assigns the highest probability of $P({\mask} = \hat{o} \mid t_r(s))$, where $t_r(s)$ represents the prompt template with the subject placeholder {\xv} replaced by $s$.
If $\hat{o}$ is the same as the gold object $o$, we conclude that the LM encodes information about the fact.
{\lama} is an evaluation benchmark, so there is no training data. It is constructed so that a pre-trained language model can be evaluated ``off-the-shelf'' with no additional fine-tuning. \citet{petroni2019language} remark that their benchmark provides only a lower-bound estimate of the amount of factual information stored in an LM, because their manually written prompts might not be optimal for eliciting facts. Accordingly, subsequent work has focused on tightening this bound by using additional training data to find more optimal prompts.

\subsection{{\lpaqa}}
\label{sec:lpaqa}
\citet{jiang2020can} use a range of text-mining and paraphrasing techniques to generate a set of candidate prompts for each relation. They collect a training dataset from Wikidata, ensuring that there is no overlap with subject-object pairs in the LAMA benchmark, and select prompts by measuring accuracy on this training data. They consider a number of rules for selecting prompts, including top-$K$ baselines and an ``optimized ensemble'', which consists of multiple prompts per relation with weights tuned on the training data. Their prompt dataset, LPAQA, is available online.\footnote{\url{https://github.com/jzbjyb/LPAQA}}

\subsection{\autop}
\label{sec:autoprompt}
\citet{shin2020autoprompt} take prompt optimization one step further by training a statistical model, {\autop}, to search over the space of input tokens for prompts that elicit correct predictions. They collect 1000 $\langle s, r, o \rangle$ triples for each relation type, either from the original T-REx dataset \cite{trex2018elsahar} or from Wikidata, with no triples that appear in the LAMA benchmark. They define a prompt for a given relation $r$ as the subject followed by a fixed number of ``trigger'' tokens:
\begin{align*}
    t_r = \xv\tv_1\tv_2\dots\tv_m\mask,
\end{align*}
where {\xv} is replaced by the subject, \tv$_i$ represents a ``trigger'' token which can be any token in the vocabulary, and the number of {\tv} tokens is set as a pre-defined number $m$.
The tokens are initialized as {\mask} tokens and then iteratively updated, at each step using a gradient-based searching algorithm~\cite{wallace2019universal} to replace one of the trigger tokens with the token that is estimated to maximize the likelihood of the gold label on the training set.

\begin{table*}[ht]
    \centering
    \begin{tabular}{l | cccc | r}
        \toprule
        \tf{Method} & {1-1} & {N-1} & {N-M} & {All} & {UHN}\\
        \midrule
        Majority & 1.8 & 23.9 & 22.0 & 22.0 & 23.8  \\
        \midrule
        {\lama} (manual) & \tf{68.0} & 32.4 & 24.7 & 31.1 &  21.8 \\
        {\lpaqa} (manual + paraphrased) & 65.0 & 35.9 & 27.9 & 34.1 & 28.7\\
        {\autop} (5 {\tv}s) & 58.0 & 46.5 & 34.0 & 42.2 & 31.3 \\
        \midrule
        {\ours} (5 \vv s) & 49.6 & 53.1 & 39.4 & 47.6 & 37.5  \\
        {\ours} (10 \vv s) & 60.7 & 53.2 & 39.2 & 48.1 &  37.9  \\
        {\ours} (manual) & 59.6 & \tf{54.1} & \tf{40.1} & \tf{48.6} & \tf{38.4}  \\
        \bottomrule
    \end{tabular}
    \caption{Micro-averaged results (top-1) on the LAMA benchmark using the BERT-base-cased model, averaged over relations. UHN stands for UnHelpfulNames~\cite{poerner2019bert}, which is a subset of LAMA where questions with helpful entity names were deleted. The LAMA results are broken down by relation category. Examples from each category are \ti{capital of} (1-1), \ti{place of birth} (N-1), and \ti{shares border with} (N-M).}
    \label{tab:lama_results}
\end{table*}


\section{Our Approach: \ours}
\label{sec:ours}
Our approach is motivated by the view that restricting the search to the space of vocabulary tokens is a suboptimal and artificial constraint.
In the case of {\autop}, optimizing over a discrete subspace is also inefficient: at each step we have to enumerate a set of candidate tokens, replace the selected trigger token, and re-run the model~\cite{shin2020autoprompt}.
The examples in Table~\ref{tab:method_comparison} also illustrate that optimized textual prompts can be opaque, despite consisting of tokens from the English vocabulary. This undermines one argument in favor of natural language prompts, which is that they are human readable so might be easier to interpret.

\paragraph{{\ours}}
In this view, we propose~{\ours}, a method for continuous prompt optimization. Rather than limiting the search to the space of discrete tokens,~{\ours} searches for optimal prompts directly, composing prompts using any vector in the embedding space.
We first follow {\autop} and define a prompt in the following form:
\begin{align*}
    t_r = \xv~\vv_1~\vv_2~\dots~\vv_m~\mask,
\end{align*}
where each {\vv}$_i \in \mathbb{R}^d$ is a dense vector with the same dimension as the LM's input embedding (e.g., $768$ for BERT-base) and the number of {\vv} vectors is set to a pre-defined number $m$.

Treating prompts as dense vectors allows us to search for optimal prompts much more efficiently.
Given some initial values for $\vv_i$, we keep all other model parameters fixed and use gradient-descent to minimize the negative log-likelihood of a training set:
\begin{equation*}
    \mathcal{L}_r = - \frac{1}{|D_r|} \sum_{(s, o) \in D_r} \log P(\mask=o \mid t_r(s)),
\end{equation*}
where $D_r$ is the set of (\ti{subject}, \ti{object}) pairs with relation $r$ and $t_r$ represents the prompt template for relation $r$ with subject tokens $s$ substituted for the placeholder {\xv}.

In this basic form, we pick a fixed value for $m$ (treated as a hyperparameter) and randomly initialize all the {\vv} tokens. We also consider a more sophisticated form of using manual prompts (we use the prompts provided in the LAMA benchmark) to decide the number as well as the \ti{position} of the {\vv} tokens for each relation and initialize each ${\vv_i}$ with the pre-trained input embedding for the corresponding tokens in the manual prompt. As shown in Table~\ref{tab:method_comparison}, we can convert a manual prompt ``{\xv} is {\mask} citizen'' into
\begin{equation*}
    t_r = {\xv} \vv_1 {\mask}  \vv_2,
\end{equation*}
and use the embeddings of \ti{is} and \ti{citizen} to initialize $\vv_1$ and $\vv_2$ respectively.  Our motivation is that a good initialization is likely to be important in this challenging non-convex optimization problem.

\paragraph{Setup}
We train {\ours} using the data collected by \newcite{shin2020autoprompt}, which contains 800 training examples with 200 held out for development.
For our main experiments, we probe the BERT-base-cased model and we compare other pre-trained language models in Appendix~\ref{appendix:compareLMs}. We report top-1 micro-averaged accuracy:\[
\frac{1}{|\mathcal{R}|} \sum_{r \in \mathcal{R}} \frac{1}{|D_r|} \sum_{(s, o) \in D_r} \mathbbm{1} [\hat{o} = o],
\] where $\mathcal{R}$ is the set of relations, $D_r$ is the set of (\ti{subject}, \ti{object}) pairs with relation $r$, and $\hat{o} = \argmax_o P(\mask = o \mid t_r(s))$. More implementation details can be found in Appendix~\ref{appendix:implementation}.

\paragraph{LAMA results} Our results are in Table~\ref{tab:lama_results}. Overall, {\ours} outperforms the previous reported results in terms of accuracy on the LAMA benchmark. Compared to {\autop}\footnote{For {\autop}, we obtain a slightly different accuracy 42.2\% by evaluating their released prompts, instead of 42.9\% reported in their paper. We suspect that this is due to a discrepancy in the vocabulary used in different papers. We use the {vocabulary} provided in the LAMA benchmark for all the evaluation: \url{https://github.com/facebookresearch/LAMA\#unified-vocabulary}.}, our models perform 5.4\%--6.4\% higher on LAMA and 6.2\%--7.1\% on the more-difficult LAMA-UHN benchmark.
The improvement is consistent across all categories, with the exception of the ``1-1'' category, which contains two relations, \ti{capital} and its inverse, \ti{capital of}.
Interestingly, the prompt that yields the best results in this category is the manual prompt, with LPAQA and {\autop} prompts performing steadily worse.
We speculate that there are very few prompts that elicit this relation with high accuracy and they are difficult to find via stochastic, non-convex optimization.

We also find that initializing the prompt vectors using the manually written prompts improves performance consistently.
This confirms our intuition that the manual initialization provides a good prior for finding a good solution in the non-convex optimization problem.
The results are broken down by relation in Table~\ref{tab:control_result_details} in the Appendix.


\section{Can We Trust Optimized Prompts?}
\label{sec:trust}
Our factual probing results confirm that {\ours} is an effective approach, outperforming the best previous method by 6.4\% on the LAMA benchmark.
However, can we conclude that BERT encodes 6.4\% more facts than was previously known?
Our prompts, like {\lpaqa} and {\autop}, are optimized on in-distribution Wikidata relations, which raises the possibility that they exploit some regularities in the underlying fact distribution.
In this section we aim to answer two questions. First, are there patterns in the Wikidata fact distribution that statistical model could theoretically exploit to predict unseen facts? Second, are optimized prompts capable of exploiting these patterns in practice?

\subsection{Facts can be predicted from training data}

\begin{table}
    \centering
    \resizebox{0.9\columnwidth}{!}{%
    \begin{tabular}{lrrrr}
        \toprule
        \tf{Relation} & \tf{Class Prior} & \tf{Naive Bayes}\\
        \midrule
All & 17.3 & 24.6 \\
1-1 & 0.2 & 0.3 \\
N-1 & 23.2 & 28.6 \\
N-M & 11.0 & 21.8 \\
\midrule
\ti{member of} & 2.2 & 59.6 \\
\ti{manufacturer} & 8.9 & 62.0 \\
        \bottomrule
    \end{tabular}
    }
    \caption{Results for simple classifiers fit to the Wikidata training data and evaluated on the LAMA test set. We highlight two relations for which object labels are correlated with particular subject tokens:
    In the \ti{member of} category, the model appears to learn that any subject with ``football'' in its name, such as \ti{Ghana Football Association}, is likely to be a member of \ti{FIFA}.
    In the \ti{manufacturer} category, the model learns to predict that \ti{Chevrolet} manufactures the \ti{Chevrolet Impala}, \ti{BMW} manufactures the \ti{BMW M Coupe}, and so on.
    }
    \label{tab:train_test_overlap}
\end{table}




We first examine whether it is possible to predict any facts by just looking at the training data.
The simplest pattern is the class prior $P(o \mid r)$: if one or two object labels dominate the relation $r$, it is easier to guess them regardless of the subject entity.
A more sophisticated pattern is to find a correlation between subject tokens and object labels---that is, to estimate $P(o \mid r, w_1, ..., w_{|s|})$, where $w_1, \ldots, w_{|s|} \in \mathcal{V}$ are the tokens of the subject name.
To see whether such patterns exist, we fit two simple probabilistic models to the Wikidata training set collected by~\citet{shin2020autoprompt}. The first model always predicts the majority class, with class priors learned from the training data, and the second is a Naive Bayes classifier (bag-of-words) with add-one smoothing (see details in Appendix~\ref{appendix:nb_implementation}). Table~\ref{tab:train_test_overlap} shows the accuracy of these models on the LAMA benchmark, averaged over relations.  The majority class model performs well because, on some relations, well over half of the examples are from the majority class.\footnote{These include \ti{native language} ($60\%$ \ti{French}) and \ti{continent} ($72\%$ \ti{Antarctica}).} The Naive Bayes baseline performs even better in all categories by learning correlations between subject tokens and object labels.
This analysis complements an observation of~\citet{poerner2019bert}, who point out that BERT can exploit superficial information in a cloze prompt to ``guess'' the correct answer---for example, predicting that people with stereotypically Italian names were likely born in \ti{Rome}.
Our results show that it is possible to learn these correlations even without prior information about entity names, and there might be other, subtler patterns in the Wikidata distribution.
\begin{figure*}
    \centering
    \includegraphics[width=0.7\paperwidth,keepaspectratio]{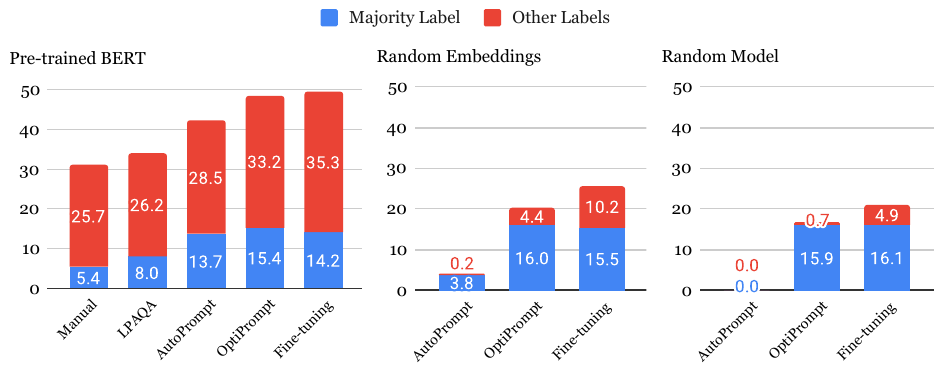}
    \caption{
    Accuracy on LAMA obtained by prompting BERT-base-cased, either the pre-trained model, reinitializing the input embeddings, or reinitializing all parameters.
    Each bar represents total accuracy micro-averaged over relations and divided into two categories: accuracy obtained by predicting the training set majority class label, and accuracy obtained by predicting other object labels. We also fine-tune BERT, which, in the random control settings, can be thought of as a better lower bound on the entropy of the task distribution.
    }
    \label{fig:controls}
\end{figure*}

\subsection{{Prompts} can exploit training data}
\label{sec:control}
We have shown that the training data clearly encodes certain regularities and simple statistical models can learn to fit the training data. In the following, we study whether a prompt optimization method built with pre-trained language models, is expressive enough to exploit these regularities in practice.
We attempt to answer this question by means of two random controls, inspired by similar proposals from linguistic probing.
In our \ti{Random Model} (RM) baseline, we optimize prompts to elicit facts from a neural network with the same architecture as the pre-trained LM but with randomly initialized parameters.
This is analogous to a \ti{control function}~\citet{pimentel2020information}, a function that removes information from a linguistic representation.
Any successful predictions in this setting must be the result of optimizing on training data.
We also consider a~\ti{Random Embeddings} (RE) baseline, where we reinitialize only the input embeddings.\footnote{In the RE setting, the classifier head of the model is also reinitialized, as the output embeddings are tied to the input embeddings.}
This is analogous to a~\ti{control task}~\cite{hewitt2019designing}, a variant of the probing task in which word types are associated with random labels.\footnote{\citet{hewitt2019designing} consider tasks like part-of-speech tagging, where each word type can be associated with a randomly selected tag. We randomize the inputs rather than the labels, which preserves most of the the statistical correlations between subject token types and object labels but removes lexical information from the embeddings.
}
Our motivation is that the Random Model setting is  more difficult to optimize, so might underestimate the ways a prompt model could exploit information from the training data.
Finally, we directly fine-tune a reinitialized BERT model on the training data with the goal of getting a better estimate of the number of LAMA facts that could be predicted from the training data.

The results are shown in Figure~\ref{fig:controls} (see implementation details and more results in Appendix~\ref{appendix:implementation} and Table~\ref{tab:control_result_details}).
In the Random Embeddings setting, both {\autop} and {\ours} are capable of finding prompts that elicit some correct predictions.
In the Random Model setting, {\autop} gets 0\% of predictions correct, presumably because it is more difficult to optimize, but {\ours} is still capable of finding successful prompts.
Most successful predictions are obtained by finding a prompt that elicits the majority class label, although {\ours} also makes a number of correct predictions that cannot be attributed to this strategy.
Our qualitative analysis suggests that these prompts exploit both class statistics and correlations between objects and subject tokens (Appendix~\ref{appendix:exploiting_training_data}).

\begin{figure*}
    \centering
    \resizebox{\linewidth}{!}{%
    \includegraphics{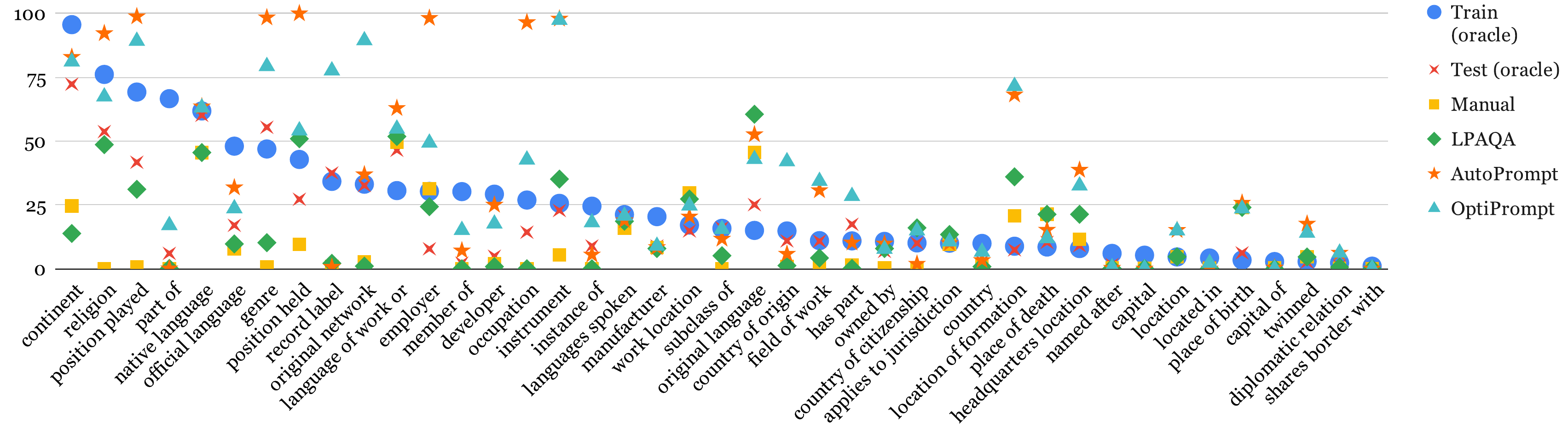}
    }
    \caption{
    The percentage of LAMA examples for which a prompt elicits the training set majority label, compared with the percentage of training and test facts with that label.
    Optimized prompts show a strong tendency to over-predict the majority class relative to manual prompts and the ground truth.
    ``Train (oracle)'' is calculated from the set of Wikidata facts collected by~\citet{shin2020autoprompt}, which is used to train {\autop} and {\ours}.
    }
    \label{fig:majority_class}
\end{figure*}

Fine-tuning BERT results in even higher accuracy, indicating that there are patterns that prompts fail to exploit.
The random controls represent a challenging setting for prompt optimization, and it is possible that the prompts are better exploiting the training data when they have access to full pre-trained BERT model.
We find evidence that this is the case by calculating how often each prompt elicits the training class majority label on LAMA, plotting the results in
Figure~\ref{fig:majority_class}. Both {\autop} and {\ours} are prone to over-predicting the majority class label.
For example, although {\autop} gets 0\% accuracy in the RM setting, it finds a prompt that elicits the majority label more than 95\% of the time for six relations when optimized on the pre-trained BERT model.\footnote{\citet{shin2020autoprompt} attempt to prevent the model from using this strategy by filtering out prompts that contain proper nouns or gold object labels, but this evidently is not enough. For example, the prompt for the \ti{position held} relation is ``{\xv} explorers voting municipal $\rightarrow$ consecrated  {\mask}.'', which elicits \ti{bishop} for 100\% of LAMA  examples.
}

LPAQA prompts predict the majority class less often, possibly because they are less effective at fitting the training distribution. However, it is still clear that LPAQA prompts also encode distribution of the training data. For instance, the highest ranked \ti{occupation} prompts discovered by LPAQA include prompts such as ``{\mask} and actors \xv'' and ``{\mask} and player \xv.'',\footnote{\url{https://github.com/jzbjyb/LPAQA/blob/master/prompt/paraphrase/P106.jsonl}} which reflect several of the most common occupations in Wikidata. We also discuss examples in Appendix~\ref{appendix:exploiting_training_data} of cases where LPAQA finds subtle changes to the prompt template that leads the model to predict the majority label more often than the manual prompt and the true test distribution. All the above evidence shows that optimized prompts can learn new facts to some extent.

\begin{table}[!t]
    \centering
    \resizebox{0.98\columnwidth}{!}{%
    \begin{tabular}{l| ccc}
        \toprule
       \tf{Method} & \tf{All} & \tf{Easy} & \tf{Hard} \\
        & (34,039) & (10,546) & (23,493)  \\
\midrule
Manual & 31.1 & 41.5 & 24.3 \\
LPAQA & 34.1 & 47.0 & 25.6 \\
{\autop} & 42.2 & 68.2 & 26.7 \\
{\ours} & \tf{48.6} & \tf{75.6} & \tf{33.0} \\
        \bottomrule
    \end{tabular}
    }
    \caption{
        Accuracy on LAMA partitioned into easy examples or hard examples, micro-averaged over relations.
        Easy facts are the facts that can be predicted by fine-tuning a BERT model with either randomly initialized parameters or randomly initialized token embeddings, or by the Naive Bayes model described in Section~\ref{sec:trust}. Hard examples are everything else.
        The numbers in parentheses denote the size of each subset.
    }
    \label{tab:lama_hard}
\end{table}

\section{How to Interpret Probing Results?}

\label{sec:lama_hard}
Our analysis in Section~\ref{sec:control} shows that optimized prompts can predict new facts from training data.
How can we interpret our factual probing results in this light?
In order to get another perspective of the relative improvement, we partition LAMA into an easy subset and a hard subset (examples from each subset can be found in Table~\ref{tab:example_predictions}).
The easy subset consists of the facts that can be correctly predicted by any of three models fit to the training data: the Naive Bayes model described in Section~\ref{sec:control} and a fine-tuned BERT model with either token embeddings reinitialized or all parameters reinitialized.
The easy subset serves as an estimate of the set of facts that can be predicted from training data.
The hard subset consists of the remain facts.
Table~\ref{tab:lama_hard} shows the results of each prompt on these two subsets of LAMA (the per-relation results are given in Table~\ref{tab:lama_hard_by_relation_type}).
First, we observe that all the probing methods achieve a much higher accuracy on the easy subset. Using more sophisticated prompt optimization techniques tends to result in big improvements on the easy subset of LAMA and smaller improvements on the hard subset.
{\ours} outperforms {\autop} by 7.4\% on the easy examples; while on the hard examples, where we filtered out facts that we know can be predicted from the training data, {\ours} also yields a big improvement (+6.3\%).
This suggests that {\ours} is both better at learning from training data and better at eliciting facts from an LM.


\begin{table*}
    \resizebox{2.1\columnwidth}{!}{%
    \begin{tabular}{ll | lllll}
        \toprule
\tf{Rel.} & \tf{Fact (manual template)} & \tf{NB} & \tf{Manual} & \tf{LPAQA} & \tf{Auto} & \tf{Opti}\\
\midrule
P103 & The native language of \ti{Jan van Krimpen} is \ti{\underline{Dutch}} . & \underline{Dutch} & \underline{Dutch} & \underline{Dutch} & \underline{Dutch} & \underline{Dutch} \\
P279 & \ti{edible mushroom} is a subclass of \ti{\underline{mushroom}} . & protein & \underline{mushroom} & category & \underline{mushroom} & \underline{mushroom} \\
P1001 & \ti{Governor of Tasmania} is a legal term in \ti{\underline{Tasmania}} . & Canada & Australia & Australia & \underline{Tasmania} & \underline{Tasmania} \\
P106 & \ti{Dude Harlino} is a \ti{\underline{actor}} by profession . & \underline{actor} & lawyer & wrestler & politician & \underline{actor} \\
P27 & \ti{Jens Evensen} is \ti{\underline{Norway}} citizen . & \underline{Norway} & Danish & Sweden & \underline{Norway} & \underline{Norway} \\
P176 & \ti{Porsche Cayenne} is produced by \ti{\underline{Porsche}} . & Honda & \underline{Porsche} & \underline{Porsche} & \underline{Porsche} & \underline{Porsche} \\
P279 & \ti{United States H-class submarine} is a subclass of \ti{\underline{submarine}} . & protein & submarines & \underline{submarine} & \underline{submarine} & \underline{submarine} \\
P138 & \ti{Milwaukee Mitchell International Airport} is named after \ti{\underline{Milwaukee}} . & Peter & Mitchell & Mitchell & \underline{Milwaukee} & \underline{Milwaukee} \\
P176 & \ti{BMW E9} is produced by \ti{\underline{BMW}} . & \underline{BMW} & \underline{BMW} & \underline{BMW} & \underline{BMW} & \underline{BMW} \\
P1412 & \ti{Tom Mann} used to communicate in \ti{\underline{English}} . & French & \underline{English} & \underline{English} & \underline{English} & \underline{English} \\
\midrule
P937 & \ti{Francis Hagerup} used to work in \ti{\underline{Oslo}} . & London & London & London & Copenhagen & \underline{Oslo} \\
P127 & \ti{Apple Store Online} is owned by \ti{\underline{Apple}} . & Germany & \underline{Apple} & \underline{Apple} & \underline{Apple} & \underline{Apple} \\
P1412 & \ti{Berengaria of Castile} used to communicate in \ti{\underline{Spanish}} . & French & \underline{Spanish} & Latin & \underline{Spanish} & \underline{Spanish} \\
P176 & \ti{SNES-CD} is produced by \ti{\underline{Sony}} . & Honda & Sega & \underline{Sony} & IBM & IBM \\
P47 & \ti{Honduras} shares border with \ti{\underline{Guatemala}} . & Lyon & \underline{Guatemala} & \underline{Guatemala} & \underline{Guatemala} & \underline{Guatemala} \\
P937 & \ti{David Ben-Gurion} used to work in \ti{\underline{Jerusalem}} . & London & \underline{Jerusalem} & \underline{Jerusalem} & \underline{Jerusalem} & \underline{Jerusalem} \\
P19 & \ti{Peter I of Serbia} was born in \ti{\underline{Belgrade}} . & Paris & \underline{Belgrade} & \underline{Belgrade} & \underline{Belgrade} & \underline{Belgrade} \\
P31 & \ti{Dally M Medal} is a \ti{\underline{award}} . & album & prize & prize & \underline{award} & \underline{award} \\
P30 & \ti{Snowdon} is located in \ti{\underline{Europe}} . & Antarctica & Wales & \underline{Europe} & Antarctica & Antarctica \\
P937 & \ti{William Lyon Mackenzie King} used to work in \ti{\underline{Ottawa}} . & London & Canada & London & Montreal & \underline{Ottawa} \\
        \bottomrule
    \end{tabular}
    }
    \caption{
     Randomly sampling 10 examples each from LAMA-easy (the first block) and LAMA-hard (the second block), only keeping examples that are predicted correctly by at least one model and that do not have the majority label.
    NB: the Naive Bayes model (Section~\ref{sec:control}), Auto: {\autop}, Opti: {\ours}.
    The correct predictions are underlined.
    }
    \label{tab:example_predictions}
\end{table*}

For a more qualitative analysis, we randomly sample ten facts from each subset, keeping only facts that are predicted correctly by at least one model and exclude examples that have the majority class label.
The examples, shown in Table~\ref{tab:example_predictions}, give a better idea of the types of predictions elicited by different prompts.
For example, both {\autop} and {\ours} appear to be exploiting the training data in some cases.
In the easy subset, they elicit more accurate predictions on cases when the answer is a token in the subject name.
In the hard subset, they show signs of having over-fit to the training distribution, incorrectly predicting the most common object labels for \ti{continent} (\ti{Antarctica}) and \ti{manufacturer} (\ti{IBM}).
{\ours} performs better than the other prompts on some facts in both categories. On an easy \ti{profession} example, while {\autop} incorrectly predicts the majority label (\ti{politician}), {\ours}---along with our Naive Bayes model---apparently encodes a lexical correlation between some aspect of the subject's name and the correct label, \ti{actor}.
On the other hand, {\ours} out-performs the other prompts on two more difficult examples: ``\ti{Francis Hagerup} used to work in \ti{Oslo}'' and ``\ti{William Lyon Mackenzie Kingused} to work in \ti{Ottawa}.'' In both cases, LPAQA predicts the training majority label (\ti{London}), {\autop} gets geographically closer (\ti{Copenhagen} and \ti{Montreal}), and {\ours} predicts the correct city.



We note that we cannot conclude that there is no way to predict these ``hard'' facts from training data.
A more general limitation of this analysis is that it does not allow us to say which strategy a model uses to make a particular prediction. Many facts can be predicted either by learning the class prior; by learning a lexical correlation between subject tokens and objects; by exploiting lexical information from the LM; or because the LM genuinely encodes information about a particular entity.
Still, the qualitative examples reveal interesting patterns in the behavior of the different prompt models that could not be observed from the summary accuracy results on the LAMA benchmark, and looking at specific predictions across a number of prompts gives us more evidence for deciding what kind of information the LM encodes about a particular fact.

\section{Discussion}
Our experiments show that {\ours} is an effective optimization algorithm, outperforming prior work at the task of eliciting facts from a pre-trained language model. However, our results are complicated by the fact that any data-driven optimization can find prompts that encode new information from the training data. This leaves open the question of which method we should select if we are interested in factual probing.

\paragraph{Continuous vs. discrete prompts}
We find that both continuous and discrete optimization are capable of finding prompts that exploit the training data.
Even when the prompt is discrete, it is rarely clear why a prompt elicits a particular prediction.\footnote{
For an illustration, see Appendix~\ref{appendix:exploiting_training_data} for a list of the {\autop} templates that elicit the majority class label more than 95\% of the time.
}
Hence, we believe that continuous prompting is more preferable, because it is easier and more efficient to optimize, and makes better predictions (in both easy and hard subsets).
On the other hand, one drawback of {\ours} (which is shared by {\autop}) is that we need white-box access to the LM to compute the gradients.
Discrete prompts will still be necessary in cases where the model parameters are not available, for example in the case of very large language models that are provided over an API.

\paragraph{Learning vs. learning to recall}
Regardless of how we choose to optimize prompts, it remains difficult to say why a model made a particular prediction---whether it was learned from training data or encoded in the LM.
Some avenues for future work might be to consider techniques for attributing predictions to specific training instances, with the goal of developing a causal understanding of how facts are acquired during pre-training or prompt optimization.
More generally, our real goal is to understand how pre-trained language models learn and represent information.
Prompt-based probing might provide some insight into this question, but we hope that future research will eventually be able to provide more mechanistic explanations for neural network behavior.
For example, it would be interesting to understand how information about entities is laid out in neural network parameters and later retrieved in response to an input prompt.

\section{Related Work}
\label{sec:related}
Our work follows from the line of factual probing experiments initiated by~\citet{petroni2019language}, who introduced the {\lama} benchmark for cloze-style factual probing. Subsequent work on {\lama} has introduced data-driven methods for optimizing prompts \cite{jiang2020can,shin2020autoprompt}.
\citet{poerner2019bert} point out that many facts in {\lama} can be predicted using lexical clues, and they introduce a new benchmark, {\lamauhn}, that is less susceptible to these heuristics.
Our work follows these projects by introducing (a) more effective techniques for optimizing prompts, and (b) a more comprehensive approach for accounting for the role of train/test overlap.
Concurrently with this work, other authors explore continuous prompt optimization:~\citet{haviv2021bertese} use an encoder to map a manually written prompt to a sequence of continuous vectors, which are then replaced with the discrete tokens that are nearby in embedding space; ~\citet{li2021prefix} propose Prefix-Tuning, which fine-tunes the left-most hidden representations in auto-regressive language models; ~\citet{liu2021gpt} use an LSTM to generate a sequence of prompt vectors.
Prompting has been explored more generally as a method for achieving ``few-shot'' learning with language models~\cite{brown2020language,schick2020size,gao2020making}.

Linguistic probing is an extensive area of research that we do not attempt to summarize here (see~\citealp{rogers2020primer} for an overview).
Our work is most related to recent proposals about how to measure whether a probe is extracting information from a representation or learning to predict the annotation from probe training data. These include random baselines \cite{hewitt2019designing} and information-theoretic measurements \cite{voita2020information}. We adopt the notion of control functions from~\citet{pimentel2020information}. Our study also relates to a larger category of work diagnosing ``shortcut learning'' \cite{geirhos2020shortcut} in neural NLP models. \citet{mccoy2019right} discover that models like BERT are often ``right for the wrong reason'', exploiting shallow heuristics rather than underlying linguistic structure, and similar effects have been discovered in many other tasks \cite{sugawara2018makes,wallace2019universal}.







\vspace{-0.3em}
\section{Conclusion}
\label{sec:conclusion}
We introduce {\ours}, an effective continuous method for optimizing prompts.
Applied to factual probing, {\ours} outperforms the best previous prompt method by 6.4\% on the {\lama} benchmark.
We find that the typical training data used for prompt optimization reveals useful information about the underlying task distribution, to the point that search algorithms can find prompts that recover ``facts'' even from a randomly initialized model.
By comparing the predictions of different prompt methods across our different controls we can form a more detailed understanding of how different prompts behave and what they can reveal about pre-trained language models.

\newpage

\section*{Ethical Considerations}
Our experiments illustrate that the ``facts'' recovered from a pre-trained language model should not be considered real facts.
Optimizing any kind of statistical model for factual prediction is likely to devolve into stereotype-learning as the model learns lexical correlations between entity names and object labels.
This problem is more pronounced if our training distribution comes from a source like Wikidata, which we find to be imbalanced.
More generally, language models that are trained on the Internet will model the toxic and harmful language that is found there, a well-documented finding for pre-trained language models like BERT~\cite[e.g.,][]{gehman2020realtoxicityprompts,nadeem2020stereoset}.
Using such models for factual prediction is liable to amplify those biases.
{\ours} is intended to be a diagnostic tool and general-purpose optimization method, not a way to use BERT as a knowledge base.

\section*{Acknowledgement}
We thank Zhengbao Jiang for answering questions about LPAQA.
We thank the members of the Princeton NLP group and the anonymous reviewers for their valuable comments and feedback. This work is supported in part by a Graduate Fellowship at Princeton University.

\bibliography{ref}
\bibliographystyle{acl_natbib}

\clearpage


\appendix


\begin{table*}[!ht]
    \centering
    \resizebox{2\columnwidth}{!}{%
    \begin{tabular}{lcccccccc}
        \toprule
         & \multicolumn{2}{c}{\tf{BERT} (110M)} & \multicolumn{2}{c}{\tf{BERT} (330M)} & \multicolumn{2}{c}{\tf{RoBERTa} (330M)} & \multicolumn{2}{c}{\tf{ALBERT} (235M)$\dagger$} \\
        \tf{Method} & Pre-T. & Rand M. & Pre-T. & Rand M. & Pre-T. & Rand M. & Pre-T. & Rand M. \\
        \midrule
        Manual & 30.6 & - & 32.2 & - & 23.6 & - & 27.4 & - \\
        \lpaqa & 35.6 & - & 36.2 & - & 29.3 & - & 29.8 & - \\
        \autop & 44.6 & 0.0 & 44.5 & 0.1 & 38.6 & 0.0 & 33.2 & 0.0 \\
        \ours & 50.8 & 19.9 & 52.7 & 19.3 & 47.8 & 19.6 & 44.6 & 16.9 \\
        \finetuning & 51.9 & 19.8 & 54.9 & 19.6 & 52.3 & 21.4 & 52.8 & 21.1 \\
        \bottomrule
    \end{tabular}
    }
    \caption{Comparison of different pre-trained LMs. We downsample the LAMA test set to make sure that in each sample, the object is a single token for all the models. $\dagger$: there is parameter sharing in ALBERT models so the actual models are much bigger. \textit{Pre-T.}: pre-trained language models. \textit{Rand M.}: randomly initialized models.
    }
    \label{tab:comparison_lms}
\end{table*}

\section{Detailed Results}

\subsection{Breakdown Accuracy for LAMA}
\label{appendix:breakdown}
Table~\ref{tab:lama_by_relation_type} shows the per-relation accuracy for each prompting method.
In many cases, we can better understand the probing results by examining the specific predictions each method makes.

\subsection{Exploiting Training Data}
\label{appendix:exploiting_training_data}

\paragraph{Majority class baseline}
Figure~\ref{fig:majority_class} shows that all optimized prompts have a tendency to over-predict the majority class label.
This behavior is most pronounced in the gradient-based methods ({\autop} and {\ours}).
It is not always clear why a particular prompt elicits these predictions.
For example,~\citet{shin2020autoprompt} attempt to prevent {\autop} from ``cheating'' by filtering out prompts that contain proper nouns or gold object labels, but there are still six relations for which {\autop} elicits the majority label more than 95\% of the time. The {\autop} prompts for these relations are:
\begin{itemize}
  \item \ti{genre} = \ti{jazz}: ``{\xv} freaking genre orchestra fiction acid {\mask}.''
  \item \ti{position played} = \ti{midfielder}: ``{\xv} played colors skier {\textbackslash}u2194 defensive {\mask}.''
  \item \ti{occupation} = \ti{politician}: ``{\xv} supporters studied politicians musician turned {\mask}.''
  \item \ti{employer} = \ti{IBM}: ``{\xv} 1987adeNBC computing succeeded {\mask}.''
  \item \ti{instrument} = \ti{piano}: ``{\xv} playingdrum concertoative electric {\mask}.''
  \item \ti{position held} = \ti{bishop}: ``{\xv} explorers voting municipal {\textbackslash}u2192 consecrated {\mask}.''
\end{itemize}
This illustrates that even discrete prompts are capable of finding prompts that elicit a specific label from an LM, and the mechanism by which these prompts elicit the prediction is often obscure.

Perhaps more surprisingly, even LPAQA occasionally finds prompts that are more likely to elicit the majority label compared to the manual prompt. The changes in these cases are often very subtle. For example, the manual prompt for the \ti{position of} relation is ``{\xv} has the position of {\mask}'' and the LPAQA prompt is ``{\xv} has the position of \tf{a} {\mask}''. Simply inserting the determiner ``a'' into the prompt leads BERT to predict the majority label, \ti{bishop}, more than five times as often compared to the manual prompt (50.9\% vs. 9.5\%), and almost twice as often relative to the true distribution in the LAMA benchmark (27.3\%).
This suggests that even simple data-driven methods can find prompts that encode some regularities in the training data and result in over-estimates of the number of facts in the language model.

\paragraph{Control result details}
Table~\ref{tab:control_result_details} shows the accuracy of optimized prompts under our random controls (Section~\ref{sec:control}) and also shows how much accuracy can be attributed to predict the majority class label.
{\autop} cannot predict any facts in the Random Model setting but performs decently on several relations in the Random Embeddings setting by predicting the majority class.
For reasons we cannot entirely explain, there is one relation, \ti{occupation}, for which {\autop}'s performance cannot be attributed to the class prior.
The correct predictions in this category are all a result of predicting \ti{actor}, which {\autop} predicts 23.3\% of the time. (The most frequent label in the training data is \ti{politician}.)
Other high frequency predictions for this relation include \ti{jet}, \ti{wool}, and \ti{smart}.
Notably, even when {\autop} finds a prompt that can draw out the class prior, it typically does not elicit the class prior 100\% of the time.

{\ours} is more successful at exploiting the training data. In the Random Model setting, virtually all correct predictions can be attributed to the majority class, which {\ours} can frequently elicit for all inputs.
One noteworthy exception is \ti{languages spoken}, where {\ours} is able to successfully classify subjects as speaking either \ti{English} or \ti{French} in some cases. It is not immediately clear what decision rule the model learns for these predictions---for example, it could be that the model predicts either \ti{English} or \ti{French} at random, in rough proportion to the training distribution; or the model is able to use the correlations between names and spoken languages.
In any case, the results illustrate that optimized prompts can learn more sophisticated strategies than simply predicting the majority class, even given a Transformer that contains no prior information at all.

\subsection{LAMA-easy and LAMA-hard}
\label{appendix:lama_hard}
Table~\ref{tab:lama_hard_by_relation_type} shows the accuracy of different prompts on the easy and hard subset of LAMA described in Section~\ref{sec:lama_hard}.
All of the optimized models tend to perform better on LAMA-easy compared to LAMA-hard,
 and {\ours} out-performs {\autop} in both categories.
For example, on the \ti{shares border} relation, {\ours} achieves an improvement on both easy questions (\ti{Campagnano di Roma}, \ti{Rome}) and hard ones (\ti{Chiapas}, \ti{Veracruz}).
But note that high accuracy on LAMA-easy does not necessarily mean that a prompt encodes information about the fact distribution. For example, all prompts, including the manually written prompts, perform well on the easy examples in the \ti{capital} relation. This category includes such facts as ``The capital of Sarajevo Canton is Sarajevo,'' which evidently do not require very much tuning to predict.

\section{Implementation Details}

\subsection{Prompt Optimization}
\label{appendix:implementation}
We implement {\ours} based on the HuggingFace's Transformers~\cite{wolf2020transformers} library.
During trianing, we use an Adam optimizer and a scheduler with a warmup ratio of $0.1$.
We use an Adam optimizer and a linear scheduler with a warmup ratio of $0.1$.
We train our {\ours} model for $10$ epochs with a learning rate of 3e-3 and a batch size of $16$.
For fine-tuning, we use an Adam optimizer and a linear scheduler with a warmup ratio of $0.1$. We fine-tune the language models for $10$ epochs with a learning rate of 2e-6 and a batch size of $2$.

We report {\autop}'s performance based on the prompts released by~\newcite{shin2020autoprompt}.
When we apply {\autop} to a control task (e.g., the Random Embeddings model), or compare {\autop} with different language models on a different dataset (see Appendix~\ref{appendix:compareLMs}), we run {\autop} for 1000 iterations for each model to search for the prompt of a relation.

\subsection{LAMA Classifiers}
\label{appendix:nb_implementation}
In Section~\ref{sec:control} we fit two simple probabilistic models to the Wikidata training data collected by~\cite{shin2020autoprompt}. Given a relation $r$ and a subject $s$ consisting of tokens $w_1, \ldots, w_{|s|} \in \mathcal{V}$, the Class Prior model predicts $\hat{o} = \argmax_o P(o \mid r)$, the object label that is most frequently associated with relation $r$ in the training data. The Naive Bayes model predicts $\hat{o} = \argmax_o P(o \mid s, r)$, with \[
P(o \mid s, r) = P(o \mid r) \prod_{i = 1}^{|s|} P(o \mid w_i).
\]
The probabilities are estimated from the corpus with add-one smoothing: \[
P(o \mid w_i) = \frac{
\mathrm{count}(o, w_i) + 1
}{
\sum_{w \in \mathcal{V}} \left ( \mathrm{count}(o, w) + 1 \right).
}
\]

\section{Comparing Pre-trained Language Models}
\label{appendix:compareLMs}
We compare different pre-trained language models (BERT, RoBERTa~\cite{liu2019roberta}, and ALBERT~\cite{lan2019albert}) with different probing methods.
We collect at most $1000$ training samples for each relation from the TRE-x dataset and constrain the object of each sample to be a token for all the models\footnote{The original data collected by \newcite{shin2020autoprompt} is not applicable when we compare different language models, because some object tokens are not in the vocabulary of RoBERTa or ALBERT.}.
During testing, we downsample the LAMA test set to make sure that the object in each sample is a single token for all the models.
In Table~\ref{tab:comparison_lms} shows the results of different probing methods applied to four pre-trained language models, along with our Random Model baseline. We make the following observations:
\begin{itemize}
    \item \textbf{Base vs. Large:} The larger version of BERT performs better on LAMA than BERT base in the {\ours} probe. We might hypothesize that BERT-large is simply more capable of finding patterns in the training data, but our baseline result does not indicate that this is the case---on the contrary, BERT-large performs marginally worse on the Random Model baseline. This could lead us to believe that BERT-large truly does store information about 1 or 2\% more LAMA facts compared to BERT-base.

    \item \textbf{BERT vs. RoBERTa vs. ALBERT:} \citet{shin2020autoprompt} find that RoBERTa performs significantly worse on LAMA than BERT. We find this is true for our prompts as well (comparing with BERT-large), but the magnitude of the difference decreases in the fine-tuning setting. Our baseline result gives a possible hint as to why: RoBERTa performs better in the RM setting with fine-tuning, indicating that part of the difference between {\ours} and fine-tuning might be due to better exploitation of training data. This change is even more dramatic in ALBERT. Perhaps these models store less factual information due to pre-training on a wider variety of genres.
\end{itemize}
We believe that further comparisons along these lines are a promising area of future work---for example, if we could show that probing results are correlated with downstream task performance and use probes to guide model selection.


\begin{table*}[ht]
    \resizebox{2.\columnwidth}{!}{%
    \begin{tabular}{lll | rrrrr}
        \toprule
        \tf{Relation} &
        \tf{Type} &
        \tf{Name} &
        \tf{\#} &
        \tf{Manual} &
        \tf{LPAQA} &
        \tf{Auto} &
        \tf{Opti} \\
\midrule

P1376 & 1-1 & capital of & 233 & \tf{73.8} & 67.8 & 56.2 & 56.7 \\
P36 & 1-1 & capital & 702 & \tf{62.1} & \tf{62.1} & 59.7 & 61.3 \\
P103 & N-1 & native language & 977 & 72.2 & 72.2 & 79.7 & \tf{86.8} \\
P127 & N-1 & owned by & 687 & 34.8 & 32.5 & 44.3 & \tf{49.6} \\
P131 & N-1 & located in the administrative territorial entity & 881 & 23.3 & 22.8 & 28.9 & \tf{41.4} \\
P136 & N-1 & genre & 931 & 0.8 & 16.8 & 55.3 & \tf{63.6} \\
P138 & N-1 & named after & 642 & 61.4 & 59.5 & 70.7 & \tf{73.4} \\
P140 & N-1 & religion & 473 & 0.6 & 59.8 & 60.5 & \tf{76.5} \\
P159 & N-1 & headquarters location & 967 & 32.4 & 35.6 & 35.7 & \tf{37.4} \\
P17 & N-1 & country & 930 & 31.3 & 39.8 & 51.0 & \tf{57.8} \\
P176 & N-1 & manufacturer & 973 & 85.5 & 81.5 & \tf{87.5} & 87.3 \\
P19 & N-1 & place of birth & 944 & \tf{21.1} & \tf{21.1} & 19.5 & 20.6 \\
P20 & N-1 & place of death & 953 & 27.9 & 27.9 & 29.8 & \tf{33.8} \\
P264 & N-1 & record label & 429 & 9.6 & 6.3 & 4.2 & \tf{45.5} \\
P276 & N-1 & location & 958 & 41.5 & 41.5 & 43.0 & \tf{47.1} \\
P279 & N-1 & subclass of & 964 & 30.7 & 14.7 & 54.9 & \tf{64.7} \\
P30 & N-1 & continent & 975 & 25.4 & 16.9 & 78.6 & \tf{86.3} \\
P361 & N-1 & part of & 932 & 23.6 & 31.4 & 37.0 & \tf{46.4} \\
P364 & N-1 & original language of film or TV show & 856 & 44.5 & 43.9 & 45.0 & \tf{51.3} \\
P37 & N-1 & official language & 966 & 54.6 & 56.8 & 52.7 & \tf{58.6} \\
P407 & N-1 & language of work or name & 877 & 64.2 & 65.2 & 68.4 & \tf{71.0} \\
P413 & N-1 & position played on team / speciality & 952 & 0.5 & 23.7 & 41.7 & \tf{44.0} \\
P449 & N-1 & original network & 880 & 20.9 & 9.1 & 33.1 & \tf{36.0} \\
P495 & N-1 & country of origin & 909 & 28.7 & 32.2 & 35.8 & \tf{40.8} \\
P740 & N-1 & location of formation & 936 & 8.9 & 13.7 & 13.1 & \tf{15.0} \\
P1001 & N-M & applies to jurisdiction & 701 & 70.5 & 72.8 & 80.5 & \tf{85.2} \\
P101 & N-M & field of work & 696 & 9.9 & 5.3 & 12.1 & \tf{14.1} \\
P106 & N-M & occupation & 958 & 0.6 & 0.0 & 13.6 & \tf{35.7} \\
P108 & N-M & employer & 383 & 6.8 & 5.7 & 7.8 & \tf{11.2} \\
P1303 & N-M & instrument & 949 & 7.6 & 18.0 & 23.1 & \tf{23.6} \\
P1412 & N-M & languages spoken, written or signed & 969 & 65.0 & 64.7 & 71.5 & \tf{76.1} \\
P178 & N-M & developer & 591 & 62.9 & 59.4 & 64.3 & \tf{67.9} \\
P190 & N-M & twinned administrative body & 992 & 2.2 & 1.7 & 2.4 & \tf{3.1} \\
P27 & N-M & country of citizenship & 966 & 0.0 & 41.5 & 45.8 & \tf{47.1} \\
P31 & N-M & instance of & 922 & 36.7 & 36.7 & 53.6 & \tf{64.9} \\
P39 & N-M & position held & 892 & 8.0 & 16.1 & 27.2 & \tf{42.8} \\
P463 & N-M & member of & 225 & \tf{67.1} & 57.3 & 64.0 & 64.0 \\
P47 & N-M & shares border with & 920 & 13.7 & 13.7 & 19.2 & \tf{22.2} \\
P527 & N-M & has part & 976 & 11.2 & 10.6 & 22.1 & \tf{34.8} \\
P530 & N-M & diplomatic relation & 996 & 2.8 & \tf{3.9} & 2.8 & 3.3 \\
P937 & N-M & work location & 954 & 29.8 & 39.1 & 34.4 & \tf{43.3} \\
\bottomrule
    \end{tabular}%
    }
    \caption{The accuracy of different prompts on LAMA for each relation using BERT-base-cased.
    {Manual:} the manually written prompts included in LAMA; {LPAQA:} manually written + paraphrased prompts from~\citet{jiang2020can}; {Auto:} the five-token {\autop} prompts released by~\citet{shin2020autoprompt}. {Opti:} {\ours} initialized using the manually written templates.
    }
    \label{tab:lama_by_relation_type}
\end{table*}


\begin{table*}
    \centering
    \resizebox{2.0\columnwidth}{!}{%
    \begin{tabular}{lll | rrr | rrr}
        \toprule
 \multirow{2}{*}{\tf{Relation}} & \multirow{2}{*}{\tf{Type}} & \multirow{2}{*}{\tf{Name}} & \multicolumn{3}{c|}{\tf{Random Embeddings}} & \multicolumn{3}{c}{\tf{Random Model}} \\
& & & Auto & Opti & FT & Auto & Opti & FT \\
\midrule
P1376 & 1-1 & capital of & 0.0/0.0 & 0.0/0.0 & 0.0/0.0 & 0.0/0.0 & 0.0/0.0 & 0.0/0.0 \\
P36 & 1-1 & capital & 0.0/0.1 & 0.0/\tf{11.8} & 0.0/0.1 & 0.0/0.0 & 0.0/0.0 & 0.0/0.0 \\
P103 & N-1 & native language & 26.5/26.5 & 60.1/60.1 & 57.6/\tf{63.5} & 0.0/0.0 & 60.1/60.1 & 60.1/\tf{60.5} \\
P127 & N-1 & owned by & 0.0/0.0 & 6.8/7.0 & 6.7/\tf{17.2} & 0.0/0.0 & 6.8/6.8 & 6.4/\tf{12.4} \\
P131 & N-1 & located in... & 0.0/0.0 & 0.0/\tf{2.3} & 0.2/1.5 & 0.0/0.0 & 0.2/0.2 & \tf{0.6}/\tf{0.6} \\
P136 & N-1 & genre & 27.6/27.6 & 55.4/55.4 & 54.7/\tf{56.1} & 0.0/0.0 & 55.3/55.3 & \tf{55.4}/\tf{55.4} \\
P138 & N-1 & named after & 0.0/0.0 & 1.2/3.9 & 0.8/\tf{50.8} & 0.0/0.0 & \tf{1.6}/\tf{1.6} & \tf{1.6}/\tf{1.6} \\
P140 & N-1 & religion & 0.0/0.0 & \tf{53.7}/\tf{53.7} & \tf{53.7}/\tf{53.7} & 0.0/0.0 & \tf{53.7}/\tf{53.7} & \tf{53.7}/\tf{53.7} \\
P159 & N-1 & headquarters location & 0.0/0.0 & \tf{9.0}/\tf{9.0} & 7.7/7.7 & 0.0/0.0 & \tf{9.0}/\tf{9.0} & \tf{9.0}/\tf{9.0} \\
P17 & N-1 & country & 0.0/0.1 & 2.8/2.8 & 1.6/\tf{7.7} & 0.0/0.0 & 0.5/1.9 & \tf{2.6}/\tf{2.6} \\
P176 & N-1 & manufacturer & 0.2/0.2 & 8.9/8.9 & 8.8/\tf{80.0} & 0.0/0.0 & 8.9/8.9 & 8.8/\tf{70.2} \\
P19 & N-1 & place of birth & 0.0/0.0 & 2.9/3.2 & 5.3/\tf{5.4} & 0.0/0.0 & 0.0/0.3 & 6.0/\tf{6.4} \\
P20 & N-1 & place of death & 0.0/0.0 & 7.3/11.4 & 7.8/\tf{12.7} & 0.0/0.0 & 10.2/10.2 & 9.7/\tf{11.6} \\
P264 & N-1 & record label & 0.0/0.0 & \tf{37.5}/\tf{37.5} & \tf{37.5}/\tf{37.5} & 0.0/0.0 & \tf{37.5}/\tf{37.5} & \tf{37.5}/\tf{37.5} \\
P276 & N-1 & location & 0.2/0.2 & 2.7/2.7 & 3.8/\tf{5.6} & 0.0/0.0 & 0.7/0.9 & 4.2/\tf{4.4} \\
P279 & N-1 & subclass of & 0.0/0.1 & 15.7/26.2 & 14.4/\tf{36.9} & 0.0/0.0 & 15.9/15.9 & \tf{16.0}/\tf{16.0} \\
P30 & N-1 & continent & 57.5/57.5 & \tf{72.3}/\tf{72.3} & \tf{72.3}/\tf{72.3} & 0.0/0.0 & \tf{72.3}/\tf{72.3} & \tf{72.3}/\tf{72.3} \\
P361 & N-1 & part of & 4.9/4.9 & 6.0/6.0 & 6.0/\tf{7.6} & 0.0/0.0 & \tf{6.0}/\tf{6.0} & \tf{6.0}/\tf{6.0} \\
P364 & N-1 & original language... & 0.0/0.0 & 21.1/25.1 & 23.5/\tf{27.8} & 0.0/0.0 & 24.1/\tf{25.5} & 22.4/\tf{25.5} \\
P37 & N-1 & official language & 0.0/0.0 & \tf{17.0}/\tf{17.0} & 14.7/16.0 & 0.0/0.0 & \tf{17.0}/\tf{17.0} & \tf{17.0}/\tf{17.0} \\
P407 & N-1 & language of work... & 0.7/0.9 & \tf{46.4}/\tf{46.4} & 45.8/\tf{46.4} & 0.0/0.0 & \tf{46.4}/\tf{46.4} & \tf{46.4}/\tf{46.4} \\
P413 & N-1 & position played... & 30.3/30.3 & 41.6/41.6 & \tf{41.7}/\tf{41.7} & 0.0/0.0 & \tf{41.7}/\tf{41.7} & \tf{41.7}/\tf{41.7} \\
P449 & N-1 & original network & 3.2/3.2 & 25.9/31.0 & 23.9/\tf{32.2} & 0.0/0.0 & 28.9/\tf{31.9} & 21.5/29.8 \\
P495 & N-1 & country of origin & 1.9/1.9 & 8.9/10.8 & 9.6/\tf{12.3} & 0.0/0.0 & 10.9/10.9 & 8.6/\tf{13.1} \\
P740 & N-1 & location of formation & 0.0/0.0 & 7.4/7.4 & 6.8/\tf{7.6} & 0.0/0.0 & 4.5/5.6 & \tf{7.4}/\tf{7.4} \\
P1001 & N-M & applies to jurisdiction & 0.1/0.1 & 8.0/42.8 & 7.4/\tf{54.9} & 0.0/0.0 & 9.6/9.6 & 9.6/\tf{9.7} \\
P101 & N-M & field of work & 0.0/0.0 & 9.6/10.1 & 10.3/\tf{10.8} & 0.0/0.0 & 10.5/10.5 & 10.5/\tf{10.8} \\
P106 & N-M & occupation & 0.0/8.9 & 6.2/27.5 & 5.2/\tf{30.8} & 0.0/0.0 & 14.3/14.3 & 6.8/\tf{26.4} \\
P108 & N-M & employer & 0.0/0.0 & 7.6/8.9 & 3.4/\tf{9.1} & 0.0/0.0 & 4.2/6.3 & 5.7/\tf{9.4} \\
P1303 & N-M & instrument & 0.0/0.0 & \tf{22.8}/\tf{22.8} & 21.9/22.7 & 0.0/0.0 & 10.6/10.6 & \tf{22.8}/\tf{22.8} \\
P1412 & N-M & languages spoken... & 0.0/0.0 & 13.9/27.7 & 10.5/\tf{28.0} & 0.0/0.0 & 7.3/25.3 & 12.0/\tf{28.2} \\
P178 & N-M & developer & 0.0/0.0 & 2.7/11.3 & 4.2/\tf{29.4} & 0.0/0.0 & 4.7/5.4 & 4.6/\tf{8.8} \\
P190 & N-M & twinned admin... & 0.0/0.0 & 0.0/1.1 & 1.7/\tf{2.1} & 0.0/0.0 & 1.4/2.1 & 0.3/\tf{2.4} \\
P27 & N-M & country of citizenship & 1.3/1.3 & 9.7/9.9 & 8.8/\tf{13.4} & 0.0/0.0 & 10.0/10.0 & 9.9/\tf{10.1} \\
P31 & N-M & instance of & 0.3/0.3 & 8.0/11.2 & 8.4/\tf{24.9} & 0.0/0.0 & 8.8/8.8 & \tf{8.9}/\tf{8.9} \\
P39 & N-M & position held & 0.0/0.0 & 20.9/28.7 & 23.8/\tf{32.4} & 0.0/0.0 & 27.2/27.2 & 18.5/\tf{30.4} \\
P463 & N-M & member of & 1.3/1.3 & 2.2/45.8 & 2.2/\tf{60.4} & 0.0/0.0 & 2.2/2.2 & 2.2/\tf{56.4} \\
P47 & N-M & shares border with & 0.0/0.0 & 0.0/0.1 & 0.2/\tf{0.9} & 0.0/0.0 & 0.0/0.0 & 0.0/\tf{1.1} \\
P527 & N-M & has part & 0.0/0.0 & 17.3/17.3 & 12.0/\tf{18.0} & 0.0/0.0 & 14.4/14.4 & \tf{17.4}/\tf{17.4} \\
P530 & N-M & diplomatic relation & 0.0/0.0 & 0.3/0.5 & 0.0/\tf{1.2} & 0.0/0.0 & \tf{0.9}/\tf{0.9} & 0.0/0.6 \\
P937 & N-M & work location & 0.0/0.0 & 14.6/14.6 & 12.1/\tf{16.8} & 0.0/0.0 & 13.2/13.2 & \tf{14.8}/\tf{14.8} \\
        \bottomrule
    \end{tabular}
    }
    \caption{
    Control result details. The value in each cell is \ti{Maj.}/\ti{Acc.}, where \ti{Acc.} is the percentage of facts of relation $r$ that the model predicts correctly and \ti{Maj.} is the percentage of facts $\langle s, r, o \rangle$ such that (a) the model predicts $o$ correctly, and (b) $o$ is the most frequent object for relation $r$ in the training data. We probe the BERT-base-cased model, reinitializing either the token embeddings or all of the parameters.
    }
    \label{tab:control_result_details}
\end{table*}


\begin{table*}[ht]
    \resizebox{2.\columnwidth}{!}{%
    \begin{tabular}{lll | rrrrr | rrrrr}
        \toprule
 \multirow{2}{*}{\tf{Relation}} & \multirow{2}{*}{\tf{Type}} & \multirow{2}{*}{\tf{Name}} & \multicolumn{5}{c|}{\tf{LAMA-easy}} & \multicolumn{5}{c}{\tf{LAMA-hard}} \\
 &  &  & \# & Man. & LPAQA & Auto & Opti & \# & Man. & LPAQA & Auto & Opti \\
\midrule
P1376 & 1-1 & capital of & 1 & \tf{100.0} & 0.0 & \tf{100.0} & 0.0 & 232 & \tf{73.7} & 68.1 & 56.0 & 56.9 \\
P36 & 1-1 & capital & 2 & \tf{50.0} & \tf{50.0} & \tf{50.0} & \tf{50.0} & 700 & \tf{62.1} & \tf{62.1} & 59.7 & 61.3 \\
P103 & N-1 & native language & 668 & 75.7 & 75.7 & 94.0 & \tf{95.7} & 309 & 64.4 & 64.4 & 48.9 & \tf{67.6} \\
P127 & N-1 & owned by & 131 & 53.4 & 87.0 & 89.3 & \tf{90.8} & 556 & 30.4 & 19.6 & 33.6 & \tf{39.9} \\
P131 & N-1 & located in... & 72 & 16.7 & 23.6 & 56.9 & \tf{63.9} & 809 & 23.9 & 22.7 & 26.5 & \tf{39.4} \\
P136 & N-1 & genre & 535 & 1.1 & 15.7 & \tf{96.1} & 92.9 & 396 & 0.3 & 18.2 & 0.3 & \tf{24.0} \\
P138 & N-1 & named after & 341 & 84.2 & 85.0 & 94.7 & \tf{95.3} & 301 & 35.5 & 30.6 & 43.5 & \tf{48.5} \\
P140 & N-1 & religion & 254 & 0.0 & 80.7 & \tf{100.0} & 96.9 & 219 & 1.4 & 35.6 & 14.6 & \tf{53.0} \\
P159 & N-1 & headquarters location & 97 & 45.4 & 57.7 & \tf{82.5} & 75.3 & 870 & 30.9 & 33.1 & 30.5 & \tf{33.2} \\
P17 & N-1 & country & 131 & 44.3 & 52.7 & 71.8 & \tf{86.3} & 799 & 29.2 & 37.7 & 47.6 & \tf{53.2} \\
P176 & N-1 & manufacturer & 791 & 94.7 & 90.8 & 95.6 & \tf{95.8} & 182 & 45.6 & 41.2 & \tf{52.2} & 50.0 \\
P19 & N-1 & place of birth & 101 & \tf{77.2} & \tf{77.2} & \tf{77.2} & 75.2 & 843 & \tf{14.4} & \tf{14.4} & 12.6 & 14.0 \\
P20 & N-1 & place of death & 183 & \tf{78.7} & \tf{78.7} & \tf{78.7} & 74.9 & 770 & 15.8 & 15.8 & 18.2 & \tf{24.0} \\
P264 & N-1 & record label & 195 & 6.2 & 6.7 & 2.6 & \tf{81.5} & 234 & 12.4 & 6.0 & 5.6 & \tf{15.4} \\
P276 & N-1 & location & 90 & 60.0 & 60.0 & 68.9 & \tf{81.1} & 868 & 39.6 & 39.6 & 40.3 & \tf{43.5} \\
P279 & N-1 & subclass of & 374 & 32.9 & 28.3 & 79.1 & \tf{92.2} & 590 & 29.3 & 6.1 & 39.5 & \tf{47.3} \\
P30 & N-1 & continent & 705 & 34.0 & 19.1 & 98.0 & \tf{99.9} & 270 & 3.0 & 11.1 & 27.8 & \tf{50.7} \\
P361 & N-1 & part of & 73 & 1.4 & 11.0 & 16.4 & \tf{98.6} & 859 & 25.5 & 33.2 & 38.8 & \tf{41.9} \\
P364 & N-1 & original language... & 373 & 69.4 & 71.8 & 76.9 & \tf{82.3} & 483 & 25.3 & 22.4 & 20.3 & \tf{27.3} \\
P37 & N-1 & official language & 197 & 43.7 & 53.8 & \tf{88.8} & 87.3 & 769 & 57.3 & \tf{57.6} & 43.4 & 51.2 \\
P407 & N-1 & language of work... & 460 & 84.6 & 85.9 & \tf{92.8} & 91.3 & 417 & 41.7 & 42.4 & 41.5 & \tf{48.7} \\
P413 & N-1 & position played... & 397 & 1.3 & 56.7 & \tf{100.0} & 99.5 & 555 & 0.0 & 0.2 & 0.0 & \tf{4.3} \\
P449 & N-1 & original network & 428 & 26.4 & 11.9 & 54.7 & \tf{64.5} & 452 & \tf{15.7} & 6.4 & 12.6 & 9.1 \\
P495 & N-1 & country of origin & 187 & 41.2 & 44.9 & 51.9 & \tf{80.2} & 722 & 25.5 & 28.9 & \tf{31.6} & 30.6 \\
P740 & N-1 & location of formation & 79 & 43.0 & 57.0 & 74.7 & \tf{86.1} & 857 & 5.7 & \tf{9.7} & 7.5 & 8.4 \\
P1001 & N-M & applies to jurisdiction & 408 & 79.9 & 84.1 & 90.7 & \tf{95.1} & 293 & 57.3 & 57.0 & 66.2 & \tf{71.3} \\
P101 & N-M & field of work & 104 & 9.6 & 12.5 & \tf{38.5} & 36.5 & 592 & 10.0 & 4.1 & 7.4 & \tf{10.1} \\
P106 & N-M & occupation & 386 & 0.0 & 0.0 & 29.8 & \tf{74.1} & 572 & 1.0 & 0.0 & 2.6 & \tf{9.8} \\
P108 & N-M & employer & 54 & 27.8 & 22.2 & 53.7 & \tf{72.2} & 329 & \tf{3.3} & 3.0 & 0.3 & 1.2 \\
P1303 & N-M & instrument & 243 & 7.8 & 37.9 & 87.7 & \tf{88.1} & 706 & 7.5 & \tf{11.2} & 0.8 & 1.4 \\
P1412 & N-M & languages spoken... & 473 & 86.7 & 81.8 & 89.4 & \tf{90.7} & 496 & 44.4 & 48.4 & 54.4 & \tf{62.1} \\
P178 & N-M & developer & 259 & 79.5 & 81.1 & 90.3 & \tf{95.4} & 332 & \tf{50.0} & 42.5 & 44.0 & 46.4 \\
P190 & N-M & twinned admin... & 44 & 2.3 & 2.3 & \tf{18.2} & 11.4 & 948 & 2.2 & 1.7 & 1.7 & \tf{2.7} \\
P27 & N-M & country of citizenship & 217 & 0.0 & 74.7 & 54.8 & \tf{75.1} & 749 & 0.0 & 31.9 & \tf{43.1} & 39.0 \\
P31 & N-M & instance of & 316 & 48.1 & 48.1 & 79.7 & \tf{94.0} & 606 & 30.7 & 30.7 & 39.9 & \tf{49.7} \\
P39 & N-M & position held & 421 & 11.2 & 28.5 & 55.3 & \tf{63.2} & 471 & 5.1 & 5.1 & 2.1 & \tf{24.6} \\
P463 & N-M & member of & 139 & 87.1 & 71.9 & 91.4 & \tf{95.0} & 86 & \tf{34.9} & 33.7 & 19.8 & 14.0 \\
P47 & N-M & shares border with & 35 & 5.7 & 5.7 & 14.3 & \tf{22.9} & 885 & 14.0 & 14.0 & 19.4 & \tf{22.1} \\
P527 & N-M & has part & 296 & 9.1 & 3.7 & 29.1 & \tf{61.8} & 680 & 12.1 & 13.5 & 19.1 & \tf{23.1} \\
P530 & N-M & diplomatic relation & 18 & \tf{5.6} & \tf{5.6} & \tf{5.6} & 0.0 & 978 & 2.8 & \tf{3.9} & 2.8 & 3.4 \\
P937 & N-M & work location & 258 & 77.1 & 86.4 & 75.6 & \tf{88.0} & 696 & 12.2 & 21.6 & 19.1 & \tf{26.7} \\
\bottomrule
    \end{tabular}%
    }
    \caption{The accuracy by relation on LAMA-easy and LAMA-hard.
    LAMA-easy consists of the facts that are predicted correctly by any of three models: the Naive Bayes model described in Section~\ref{sec:trust}; BERT-base-cased with randomly initialized token embeddings; and BERT-base-cased with all parameters reinitialized.
    LAMA-hard contains all the remaining facts.}
    \label{tab:lama_hard_by_relation_type}
\end{table*}


\end{document}